\documentclass[11pt]{article}
\usepackage[margin=1in]{geometry}
\usepackage{amsmath}
\usepackage{amssymb}
\usepackage{amsfonts}
\usepackage{mathtools}
\usepackage{graphicx}
\usepackage{url}
\usepackage{hyperref}

\title{\textbf{An Equal-Probability Partition of the Sample Space: \\ A Non-parametric Inference from Finite Samples}}
\author{Urban Eriksson \\ \texttt{123ube@gmail.com} \\ \textit{Independent Researcher}}
\date{\today}

\begin{document}

\maketitle

\begin{abstract}
This paper investigates what can be inferred about an arbitrary continuous probability distribution from a finite sample of $N$ observations drawn from it. The central finding is that the $N$ sorted sample points partition the real line into $N+1$ segments, each carrying an expected probability mass of exactly $1/(N+1)$. This non-parametric result, which follows from fundamental properties of order statistics, holds regardless of the underlying distribution's shape. This equal-probability partition yields a discrete entropy of $\log_2(N+1)$ bits, which quantifies the information gained from the sample and contrasts with Shannon's results for continuous variables. I compare this partition-based framework to the conventional ECDF and discuss its implications for robust non-parametric inference, particularly in density and tail estimation.
\end{abstract}

\section{Introduction}

Inferring the properties of a probability distribution from a finite sample is a fundamental challenge in statistics and information theory. The empirical cumulative distribution function (ECDF) is a well-known non-parametric estimator of the true distribution function $F(x)$. Given $N$ independent and identically distributed observations, $x_{(1)} \le x_{(2)} \le \cdots \le x_{(N)}$ (the sorted sample), the ECDF is defined as a step function that increases by $1/N$ at each observed data point. In other words, it is defined as $\hat{F}_N(x) = \frac{1}{N}\sum_{i=1}^N \mathbf{1}_{\{X_i \le x\}}$, assigning a probability mass of $1/N$ to each observation. Thus $\hat{F}_N(x_{(i)})$ equals $i/N$ for the $i$-th order statistic, with the function reaching 1 at the largest observation, $x_{(N)}$. The ECDF is a consistent estimator of $F(x)$ and is unbiased for any fixed $x$. However, the ECDF's discrete nature, placing all probability mass at the sample points, may not fully capture the uncertainty about regions where no observations have fallen, especially in the tails beyond the sample extremes. Moreover, it provides no natural framework for reasoning about the probability content of intervals.

This paper presents an alternative non-parametric framework based on the principle of an equal-probability partition of the sample space. Instead of assigning probability to the points themselves, this approach assigns probability to the intervals they define. This resembles constructing a variable-width histogram where each bin contains equal probability. I demonstrate that when the $N$ sorted observations partition the real line into $N+1$ segments (including the two semi-infinite tails), each segment carries an \textit{expected} probability mass of exactly $1/(N+1)$. Remarkably, this implies that the unobserved tail regions have the same expected probability as any interval between adjacent observations. Although initially counter-intuitive, this result is a natural consequence of symmetry and the absence of prior knowledge about the distribution's shape.

This equal-probability partition effectively spreads out probability mass uniformly across the $N+1$ intervals, as defined by the sample. The approach embodies the principle of insufficient reason in the absence of evidence favoring one region over another, treating all intervals as equally likely. This framework is particularly valuable for practitioners dealing with small samples, extreme value analysis, or situations requiring transparent uncertainty quantification. It provides a principled approach when traditional methods may overstate confidence in regions beyond the observed data. Furthermore, this perspective provides a new interpretation of established results in order statistics. The well-known result that $E[F(x_{(i)})] = i/(N+1)$~\cite{mood1974,morrison1976} follows naturally from the presented framework. It represents the cumulative probability up to the $i$-th segment boundary, having accounted for $i$ segments each with expected probability $1/(N+1)$. Section 2 presents three complementary derivations of this equal-probability partition result. Section 3 discusses its implications for probabilistic interpretation, plotting positions, and information theory. Section 4 examines the framework's connections to existing statistical methods and its advantages for tail estimation.

\section{Derivations of the Equal-Probability Partition}

\subsection{Setup}

Consider a continuous random variable $X$ with unknown cumulative distribution function (CDF) $F(x)$ and probability density function (PDF) $f(x)$ on $\mathbb{R}$. Given a sample of $N$ independent and identically distributed observations, $X_1, X_2, \dots, X_N$, let $x_{(1)} < x_{(2)} < \cdots < x_{(N)}$ denote the corresponding order statistics. These order statistics partition the real line into $N+1$ disjoint segments:

\begin{itemize}
    \item Segment 0: $(-\infty, x_{(1)})$, the region below the smallest observation.
    \item Segment $i$: $(x_{(i)}, x_{(i+1)})$ for $i=1, 2, \dots, N-1$, the intervals between successive observations.
    \item Segment $N$: $(x_{(N)}, +\infty)$, the region above the largest observation.
\end{itemize}

Let $P_i$ denote the probability mass that the true distribution $F$ assigns to the $i$-th segment. Since the locations of the sample points $x_{(i)}$ are random, these segment probabilities $P_i$ are themselves random variables. The objective is to determine the expected value of these probabilities, $E[P_i]$, given only the sample size $N$ and without assumptions about $F$.

These random variables $P_i$ are expressed in terms of the true (but unknown) CDF as follows:
\begin{align*}
    P_0 &= F(x_{(1)}) \\
    P_i &= F(x_{(i+1)}) - F(x_{(i)}) \quad \text{for } 1 \le i \le N-1 \\
    P_N &= 1 - F(x_{(N)})
\end{align*}
By construction, these probabilities form a complete partition: $\sum_{i=0}^N P_i = 1$.

\subsection{Derivation from the First Order Statistic}

As a first approach, this section proves that the expected probability mass in the first segment, $E[P_0]$, equals $1/(N+1)$. The proof uses the distribution of the first order statistic, $x_{(1)}$.

By definition, we have $P_0 = F(x_{(1)})$. To compute its expected value, we first need the probability distribution of the random variable $x_{(1)}$. The CDF of $x_{(1)}$ follows from noting that  $\{x_{(1)} \le x\}$ occurs if and only if at least one of the $N$ samples is at most $x$, which is the complement of the event that all $N$ independent samples exceed $x$. The probability of a single sample being greater than $x$ is $1-F(x)$. Therefore, the CDF of $x_{(1)}$ is:
\begin{equation}
    F_{x_{(1)}}(x) = P(x_{(1)} \le x) = 1 - P(\text{all } N \text{ samples} > x) = 1 - [1-F(x)]^N
\end{equation}
Differentiating this with respect to $x$ yields the PDF of $x_{(1)}$:
\begin{equation}
    f_{x_{(1)}}(x) = \frac{d}{dx}F_{x_{(1)}}(x) = N[1-F(x)]^{N-1}f(x)
\end{equation}
The expected value of $P_0 = F(x_{(1)})$ is computed by integrating $F(x)$ against the PDF of $x_{(1)}$:
\begin{equation}
    E[P_0] = E[F(x_{(1)})] = \int_{-\infty}^{+\infty} F(x) \, f_{x_{(1)}}(x) \, dx
\end{equation}
Substituting the PDF $f_{x_{(1)}}(x)$ and applying a change of variables to $u = F(x)$ (so that $du = f(x)dx$), the integral transforms as follows:
\begin{equation}
    E[P_0] = \int_{-\infty}^{\infty} F(x) N[1-F(x)]^{N-1} f(x) dx = N \int_0^1 u [1-u]^{N-1} du
\end{equation}
The integral on the right is a Beta integral with parameters $a=2$ and $b=N$. Using $B(a,b)=\frac{\Gamma(a)\Gamma(b)}{\Gamma(a+b)}$, we have:
\begin{equation}
    E[P_0] = N \cdot \left( \frac{1}{N(N+1)} \right) = \frac{1}{N+1}
\end{equation}
This elegant result shows that, regardless of the underlying distribution $F$, the \textbf{expected} cumulative probability below the first order statistic is exactly $1/(N+1)$. This provides the foundation for the general result, which can be established by extending this logic to all segments.

\subsection{Derivation from Symmetry and Uniform Spacings}

A more general derivation uses the principle of symmetry, and can be understood by looking at the probability integral transform (PIT). This approach proves the result for all $N+1$ segments simultaneously.

Consider transforming the original random variables $X_i$ via their own (unknown) CDF, $F$. Let $U_i = F(X_i)$. By applying the probability integral transform the $U_i$ become independent and identically distributed as Uniform(0,1). The order statistics of the original sample, $x_{(1)}, \dots, x_{(N)}$, map directly to the order statistics of this uniform sample, denoted $u_{(1)} < u_{(2)} < \cdots < u_{(N)}$, where $u_{(i)} = F(x_{(i)})$.

These uniform order statistics partition the interval $[0,1]$ into $N+1$ random sub-intervals, known as spacings. Let the lengths of these spacings be denoted by $\Delta_i$:
\begin{align*}
    \Delta_0 &= u_{(1)} \\
    \Delta_i &= u_{(i+1)} - u_{(i)} \quad \text{for } 1 \le i \le N-1 \\
    \Delta_N &= 1 - u_{(N)}
\end{align*}
It is well-known that the joint distribution of this vector of spacings $(\Delta_0, \Delta_1, \dots, \Delta_N)$ is the Dirichlet distribution with parameters $\alpha_0 = \alpha_1 = \cdots = \alpha_N = 1$~\cite{david2003}. This specific case is equivalent to a uniform distribution over the standard simplex in $\mathbb{R}^{N+1}$.

The key property of the Dirichlet($1, 1, \dots, 1$) distribution is its complete symmetry. The expected value for any component $\Delta_i$ is given by:
\begin{equation}
    E[\Delta_i] = \frac{\alpha_i}{\sum_{j=0}^N \alpha_j} = \frac{1}{N+1}
\end{equation}
Note that these spacings $\Delta_i$ coincide with the segment probabilities $P_i$ from Section 2.1. For example, $P_0 = F(x_{(1)}) = u_{(1)} = \Delta_0$, and $P_i = F(x_{(i+1)}) - F(x_{(i)}) = u_{(i+1)} - u_{(i)} = \Delta_i$. Therefore, by the symmetry of the uniform spacings we have that:
\begin{equation}
    E[P_i] = E[\Delta_i] = \frac{1}{N+1} \quad \text{for all } i = 0, 1, \dots, N
\end{equation}
This confirms and generalizes the result from the previous section, providing a unified justification for the equal-expected-probability partition.

\subsection{Derivation from Induction}

The third and final derivation confirms the result through induction. The approach is to solve for the expected probability of each segment sequentially, treating each as the first segment of a progressively smaller conditional problem.

The recursion begins with the result from Section 2.2: for a problem with $k$ sample points, the expected probability mass in the first segment (below the lowest point) is $1/(k+1)$.

\textbf{Step 0: Derive $E[P_0]$} \\
For our main problem with $N$ points, we apply this rule directly to find the expected mass in the first segment, $P_0$:
\begin{equation}
    E[P_0] = \frac{1}{N+1}
\end{equation}

\textbf{Step 1: Derive $E[P_1]$} \\
Consider the conditional distribution given $X > x_{(1)}$. The total expected probability mass available in this space is $E[1 - P_0] = 1 - 1/(N+1) = N/(N+1)$. This conditional distribution involves the remaining $N-1$ sample points, $x_{(2)}, \dots, x_{(N)}$. The segment $P_1$ (the interval $(x_{(1)}, x_{(2)})$) is the \textit{first segment} of this new conditional problem.

Applying the result from Section 2.1 to this collection of $N-1$ points, the expected share of probability that $P_1$ receives \textit{from the conditional mass} is $1/((N-1)+1) = 1/N$. To find the unconditional expectation, we multiply this share by the total expected mass available in the conditional space:
\begin{equation}
    E[P_1] = \underbrace{\left(\frac{1}{N}\right)}_{\text{Share of conditional mass}} \times \underbrace{\left(\frac{N}{N+1}\right)}_{\text{Total conditional mass}} = \frac{1}{N+1}
\end{equation}

\textbf{Step $i$: The General Recursive Step} \\
Assuming by induction that for all segments up to $k = i - 1$, $E[P_k] = 1/(N+1)$, we now derive $E[P_i]$. We define a new conditional problem space for $X > x_{(i)}$. The total expected probability mass available is:
\begin{equation*}
    1 - \sum_{k=0}^{i-1} E[P_k] = 1 - \frac{i}{N+1} = \frac{N+1-i}{N+1}
\end{equation*}
This problem contains the remaining $N-i$ sample points, and our segment $P_i$ is its first segment. Applying our rule, the expected share $P_i$ receives from this conditional mass is $1/((N-i)+1)$. Its unconditional expectation is therefore:
\begin{equation}
    E[P_i] = \underbrace{\left(\frac{1}{N-i+1}\right)}_{\text{Share}} \times \underbrace{\left(\frac{N+1-i}{N+1}\right)}_{\text{Mass}} = \frac{1}{N+1}
\end{equation}
This confirms the result for all $i < N$. For the final tail segment, $P_N$, its expected mass is simply the remainder of the total probability: $E[P_N] = 1 - \sum_{k=0}^{N-1} E[P_k] = 1 - N/(N+1) = 1/(N+1)$. This completes the inductive proof.

\textit{It is worth emphasizing that all three derivations establish a result that is distribution-free; it does not depend on any particular parametric form of $F(x)$. It holds true for any continuous i.i.d. sample, relying only on the fundamental properties of order statistics and the absence of any prior bias favoring a specific region of the support.}

\section{Results: Distribution and Entropy Implications}

\subsection{Probabilistic Interpretation and Plotting Positions}

The result that $E[P_i] = 1/(N+1)$ for all segments has profound implications. In the absence of additional information, the most rational non-parametric model assigns equal probability to each sample-defined segment. This aligns with the principle of maximum entropy: for a discrete set of $N+1$ possible outcomes (i.e., which segment a new data point will fall into), the probability distribution with the highest entropy is the uniform one, where each outcome has a probability of $1/(N+1)$. This represents the state of maximum uncertainty, or minimum prior assumption, consistent with the partition.

From a Bayesian perspective, this corresponds to using a non-informative Dirichlet$(1,1,\dots,1)$ prior for the vector of segment probabilities $(P_0, P_1, \dots, P_N)$. As noted in Section 2.3, this distribution is uniform over the simplex. Since the specific locations of the $x_{(i)}$ do not provide information to alter our belief about the relative sizes of these probability spacings, this prior remains the posterior. The expected value for each segment probability under this distribution is precisely $1/(N+1)$, confirming the earlier result.

This framework has direct applications to quantile estimation and probability plotting. An empirical estimate of the CDF based on this partition, denoted $\tilde{F}(x)$, assigns to the $i$-th order statistic the sum of the expected probabilities in the segments up to that point:
\begin{equation}
    \tilde{F}(x_{(i)}) = \sum_{j=0}^{i-1} E[P_j] = \frac{i}{N+1}
\end{equation}
This coincides with the classical result that  $E[F(x_{(i)})] = i/(N+1)$, which follows from $F(x_{(i)})$ having a Beta$(i, N-i+1)$ distribution~\cite{mood1974}. The quantities $i/(N+1)$ are widely used as plotting positions in practice. The equal-probability partition thus provides a fundamental justification for these widely used plotting positions, showing they arise naturally from first principles. This approach also avoids the conceptual difficulty of the ECDF, which assigns cumulative probability $1$ to $x_{(N)}$, implying no values can exceed the sample maximum.

\subsection{Comparison to the Traditional ECDF}

The CDF estimate based on the equal-probability partition, $\tilde{F}(x)$, and the classical ECDF, $\hat{F}_N(x)$, differ fundamentally in how they allocate probability mass. The ECDF places a point mass of $1/N$ at each observed value and its value jumps to 1 at the largest observation, $x_{(N)}$. In contrast, the partition-based approach assigns no probability mass to the points themselves, but instead assigns an expected probability of $1/(N+1)$ to each interval. This difference is most pronounced in the tails:
\begin{itemize}
    \item At the first order statistic, $\hat{F}_N(x_{(1)}) = 1/N$ while $\tilde{F}(x_{(1)}) = 1/(N+1)$.
    \item At the last order statistic, $\hat{F}_N(x_{(N)}) = 1$ while $\tilde{F}(x_{(N)}) = N/(N+1)$.
\end{itemize}
The partition-based approach thus inherently reserves an expected probability mass of $1/(N+1)$ for values beyond the observed maximum, explicitly acknowledging the uncertainty about the upper tail of the distribution.

The difference becomes clearer when viewed in terms of density. The ECDF corresponds to an empirical probability mass function concentrated entirely at the sample points, often visualized as a sum of Dirac delta functions of weight $1/N$ at each $x_{(i)}$. By contrast, a natural density estimator can be constructed from the equal-probability partition by distributing the mass $1/(N+1)$ uniformly over each respective segment. This creates a piecewise-uniform density function, effectively a histogram with variable bin widths determined by the data. The bin widths equal the gaps between the data points, with the height of each bar set to ensure an area of $1/(N+1)$. Consequently, wide gaps between observations (suggesting low density) result in short histogram bars, while tight clusters of observations (suggesting high density) result in tall bars.

This constructed density should not be interpreted as the ``true'' density. Rather, it represents a maximally non-informative estimator, consistent with the constraints imposed by the observed order statistics. For small $N$, equal segment probabilities provide a coarse approximation. As $N$ grows and the sample points better resolve the distribution's support, this estimator converges to the true density. In the limit, as $N \to \infty$, the difference between fractions with $N$ and $N+1$ in the denominator becomes negligible. Both $\tilde{F}(x)$ and $\hat{F}_N(x)$ are consistent estimators that converge to the true CDF, $F(x)$. The primary advantage of the equal probability partition is for finite samples, where it provides a correction that explicitly accounts for uncertainty, particularly in the tails.

\subsection{Information Content and Entropy of the Partition}

The equal-probability partition enables a natural quantification of the information it contains using Shannon entropy. The partition defines a complete set of $N+1$ mutually exclusive outcomes for the location of a new observation. Since each outcome has probability $1/(N+1)$, the entropy $H$ is:
\begin{equation}
    H = -\sum_{i=0}^N \frac{1}{N+1} \log_2\left(\frac{1}{N+1}\right) = -(N+1) \left(\frac{-\log_2(N+1)}{N+1}\right) = \log_2(N+1) \text{ bits}
\end{equation}
This quantity measures the information content, or equivalently the uncertainty, in predicting which segment will contain a future observation, given $N$ samples. For example, with $N=3$ observations partitioning the space into 4 segments, the entropy is $H=\log_2(4)=2$. With $N=9$, it is $H=\log_2(10) \approx 3.32$ bits.

The growth of this entropy is sublinear, with each additional sample providing a diminishing informational return. Also, note that the argument of the logarithm is $N+1$ rather than a naive $N$. This is a direct consequence of the partition structure, which accounts for the two tail segments in addition to the $N-1$ inter-sample segments. The presence of these tails is a crucial feature of the model, reflecting the fact that a finite sample cannot eliminate uncertainty about values beyond its observed range. Thus $H=\log_2(N+1)$ represents the theoretical minimum number of bits required to encode the location of a new sample relative to the existing partition.

This result stands in contrast to the classical work of Shannon~\cite{shannon1948} on continuous variables. A non-degenerate continuous random variable is understood to have infinite entropy, as specifying a single draw with arbitrary precision requires an infinite number of bits. This framework sidesteps this issue because the approach does not model the raw continuum; instead it models a \textit{finite partition induced by the data}. By discretizing the problem into $N+1$ possible outcomes, I arrive at a finite and highly practical measure of uncertainty. This approach reconciles with classical theory in the limit: as $N \to \infty$, the partition becomes infinitely fine, and the entropy $H = \log_2(N+1)$ diverges to infinity, as expected for a continuous variable. For any real-world inference, however, we must operate with a finite $N$. The entropy $H=\log_2(N+1)$ thus provides a meaningful quantification of the information extractable from a finite sample.

\section{Discussion}

The equal-probability partition framework is a foundational non-parametric estimation technique with connections to established methods like histogram and kernel density estimation. Specifically, a density estimator built on this partition is a form of an \textit{equal-frequency} histogram. While a standard histogram has fixed-width bins with variable counts, an equal-frequency histogram has variable-width bins, each containing (approximately) the same number of observations. This model represents the most extreme case, where the bin boundaries are the order statistics themselves, and thus each interior bin contains zero observations by construction. This perspective reveals the model's role as a fundamental, unsmoothed building block for density estimation. One could use this partition as a starting point for further refinement. For example, applying a kernel to the uniform density within each segment while preserving the total segment mass of $1/(N+1)$ would generate a smoother estimate that still respects the equal-probability principle. The partition-based piecewise-uniform density itself is the maximally conservative choice, since it does not speculate about bumps or trends within segments and adheres strictly to the information provided by the order statistics.

From a Bayesian perspective, my approach is equivalent to placing a non-informative, symmetric Dirichlet$(1,1,\dots,1)$ prior on the probability vector $(P_0, P_1, \dots, P_N)$ corresponding to the $N+1$ segments. Since the data consist only of the locations of the boundaries, and no observations fall \textit{within} the interior bins, this prior also serves as the posterior. In contrast, the classical ECDF is philosophically closer to models like the Dirichlet Process or the Bayesian bootstrap, which place discrete probability mass directly at the observed data points. The distinction is philosophically important: the ECDF implicitly assumes the observed values represent the only locations of support, whereas the partition framework treats them merely as boundaries between regions of equal expected probability.

A key advantage of this framework is its treatment of tail estimation. By explicitly assigning an expected probability mass of $E[P_0] = E[P_N] = 1/(N+1)$ to the two semi-infinite tail regions, the model formally acknowledges uncertainty beyond the observed sample range. This is particularly important for heavy-tailed distributions, where the largest observation, $x_{(N)}$, may be a significant underestimate of the distribution's true extent. In contrast, the ECDF, which places all probability at or below $x_{(N)}$, can severely underestimate tail risk. The partition-based model, by contrast, allocates a non-zero mass beyond $x_{(N)}$, offering a more realistic assessment, especially with small samples. Similarly, the non-zero probability below $x_{(1)}$ provides a buffer that can be critical if the distribution has a lower bound that was not captured by the sample.

The framework is also deeply connected to the formal theory of order statistics. As mentioned, the result that $E[F(x_{(i)})] = i/(N+1)$ is a classical one, derived from the fact that $F(x_{(i)})$ follows a Beta$(i, N-i+1)$ distribution. The model therefore constructs a CDF estimator, $\tilde{F}(x)$, using the \textit{mean} of the underlying Beta distribution as the point estimate for the cumulative probability at each order statistic. This choice is sensible as it is unbiased on average. While alternative estimators could be built using other properties of the Beta distribution (e.g., the median, which leads to plotting positions like $(i-0.3)/(N+0.4)$), using the mean provides the elegant symmetry that leads directly to the simple, closed-form entropy of $\log_2(N+1)$. For a comprehensive review of different plotting position formulas used in practice, see Hyndman and Fan~\cite{hyndman1996}.

A crucial distinction is that while the model yields equal probability segments, the interval lengths, $|x_{(i+1)}-x_{(i)}|$ are not equal. When the sample contains tight clusters of points, the corresponding intervals will be narrow. Since each still contains an expected mass of $1/(N+1)$, the implied density (probability per unit length) in these regions will be high. Conversely, large gaps between observations will form wide intervals, implying a low local density. This behavior qualitatively matches our intuition about the underlying distribution. While this approach exaggerates the effect by enforcing exact equality of probability mass across segments, it provides a neutral, data-driven starting point when no additional information is available.

\section{Conclusion}

This paper has presented a non-parametric framework based on the principle of an equal-probability partition. By leveraging the distribution-free properties of order statistics, I have shown that for a sample of size $N$, the $N+1$ segments defined by the order statistics each carry an expected probability mass of exactly $1/(N+1)$. This finding leads to a natural, piecewise-uniform estimate of the underlying density. This contrasts with the traditional ECDF, which places discrete point masses at the observations. The resulting partition has a discrete entropy of $\log_2(N+1)$ bits, a finite measure that encapsulates the information gained from a finite sample and highlights the role of data-induced discretization in statistical inference.

The partition-based approach provides a simple, intuitive tool for understanding what can (and cannot) be known about a distribution from limited data. It emphasizes the uncertainty in unobserved regions by guaranteeing them a fair share of the total probability mass. While it may not replace more sophisticated density estimators for all practical tasks, it provides a fundamental baseline that is more transparent about distributional uncertainty than the empirical step function. Moreover, this framework provides a theoretical foundation for established techniques like plotting positions in quantile estimation and can inform adaptive histogram binning.

Future research might extend this framework by incorporating prior knowledge or applying smoothing techniques within segments. Quantifying the variability of the segment probabilities $P_i$, around their expected value of $1/(N+1)$ for finite samples would also be a valuable contribution. Furthermore, extending the concept to multidimensional data, for example by partitioning space into polyhedral cells of equal expected probability, presents an interesting and challenging avenue with connections to computational geometry and optimal transport.

Ultimately, the equal-probability partition offers a theoretically grounded way to infer a plausible distribution from $N$ points under minimal assumptions. Its core directive is simple: treat each sample-defined interval as equally likely. It serves as a powerful reminder of the principled ignorance one should maintain when faced with limited information, an insight that is both intuitively appealing and rigorously supported by order statistics and information theory.

\end{document}